\documentclass[submission,copyright,creativecommons]{eptcs}
\providecommand{\event}{SOS 2007} 
\usepackage{breakurl}             
\usepackage{cite}
\usepackage{amsmath,amssymb,amsfonts}
\usepackage{algorithmic}
\usepackage{graphicx}
\usepackage{textcomp}
\usepackage{xcolor}
\usepackage{multirow}
\usepackage{array}
\usepackage{float}
\usepackage{subcaption}
\usepackage{capt-of}
\usepackage{epstopdf}
\AppendGraphicsExtensions{.pdf}

\begin{document}

\title{Model Compression for Resource-Constrained\\ Mobile Robots}

\author{Timotheos Souroulla
\institute{Ericsson Research AI}
\institute{Ericsson AB, Sweden\\
Stockholm, Sweden}
\email{timotheos.souroulla@ericsson.com}
\and
Alberto Hata
\institute{Ericsson Research AI}
\institute{Ericsson Telecomunica\c{c}\~{o}es S/A \\
Indaiatuba, Brazil}
\email{alberto.hata@ericsson.com}
\and
Ahmad Terra
\institute{Ericsson Research AI}
\institute{Ericsson AB \\
Stockholm, Sweden}
\email{ahmad.terra@ericsson.com}
\and
\"Ozer \"Ozkahraman
\institute{Robotics Perception and Learning}
\institute{KTH, Royal Institute of Technology \\
Stockholm, Sweden}
\email{ozero@kth.se}
\and
Rafia Inam
\institute{Ericsson Research AI}
\institute{Ericsson AB \\
Stockholm, Sweden}
\email{rafia.inam@ericsson.com}}

\def\titlerunning{Model Compression for Resource-Constrained Mobile Robots}
\def\authorrunning{T. Souroulla, A. Hata, A. Terra, \"O. \"Ozkahraman \& R. Inam}
\maketitle 

\begin{abstract}
The number of mobile robots with constrained computing resources that need to execute complex machine learning models has been increasing during the past decade.
Commonly, these robots rely on edge infrastructure accessible over wireless communication to execute heavy computational complex tasks. However, the edge might become unavailable and, consequently, oblige the execution of the tasks on the robot. This work focuses on making it possible to execute the tasks on the robots by reducing the complexity and the total number of parameters of pre-trained computer vision models. This is achieved by using model compression techniques such as Pruning and Knowledge Distillation. These compression techniques have strong theoretical and practical foundations, but their combined usage has not been widely explored in the literature. Therefore, this work especially focuses on investigating the effects of combining these two compression techniques. The results of this work reveal that up to $90\%$ of the total number of parameters of a computer vision model can be removed without any considerable reduction in the model's accuracy.
\end{abstract}

\section{Introduction}

The amount of scenarios where robots (which are mostly resource-constrained) that need to execute complex machine learning models is increasing rapidly. An example of such a scenario is a Human-Robot Collaboration (HRC) scenario where robots need to avoid any hazardous situations through safety analysis \cite{inam2018safety, 8502466}. Commonly, safety analysis involves complex computer vision tasks that require substantial computing power to perform the inferences in real-time.

Many robots have constrained computing resources that might not fulfill the requirement of executing the safety analysis module without delays, which may lead to an increase in the risk level. Therefore, they rely on edge infrastructures to run these complex tasks. In some cases, the communication between the robot and the edge is not possible and in other cases, a less accurate inference result is allowed in low-risk situations (e.g. absence of nearby obstacles). 
These situations motivate devising solutions to process the robot's sensor data on its embedded processor rather than executing on the edge. 

Through the years, different approaches aiming to tackle computer vision and object detection tasks were investigated, leading to a lot of different computer vision models. Most of these algorithms have the major disadvantage of being highly complex, thus they require a lot of processing and computing power to be executed. Consequently, mobile robots with constrained computing resources cannot execute them on their processors without any delays. Therefore, an investigation seeking to reduce the original version’s complexity and the required computational load will be beneficial.

This work proposes reducing the complexity of the computer vision models through model compression to allow their execution on the robot's embedded computer. Essentially, model compression reduces the complexity of a model by removing network connections that have low contribution in the output generation. In addition, computer vision models can have a plethora of redundant operations that do not need to be processed, such as multiplications by $1$ or additions with $0$. Two techniques commonly employed in model compression problems are evaluated: Pruning \cite{marchisio2019deep}, which identifies and removes redundant operations, and Knowledge Distillation \cite{matsubara2020head}, which replaces the original model with a smaller, less complex one that mimics the output of the original one. Given these techniques, it is verified whether the compressed models can maintain high levels of accuracy while reducing their total number of parameters. 
The solution is aimed to be employed in safety-critical HRC scenarios where real-time requirements should be satisfied along with accurate responses of the models.

The main contribution of this work is a model compression solution that combines different strategies to reduce the complexity of computer vision models with low degradation in accuracy. More specifically, Knowledge Distillation and Pruning are combined which showed better performance compared to their stand-alone versions.

The rest of this paper is structured as follows: \autoref{ch:background} presents related work of model compression techniques; \autoref{ch:methods} presents the methods used for compressing a deep neural network (DNN) model; \autoref{ch:experiments} presents the experimental setup; \autoref{ch:resultsAndAnalysis} provides an analysis and comparisons between the results; Sections \ref{ch:discussion} and \ref{ch:conclusionsAndFutureWork} conclude this work with discussions and future works.

\section{Related Works}
\label{ch:background}
The reduction of the complexity and the computational load of complex computer vision tasks is done for multiple reasons. 
One of them is the constrained computing resources of mobile robots which prohibits real-time execution of these tasks. As mentioned in \cite{marchisio2019deep}, the most important reason for compressing a model is the low power and memory budget of a mobile robot as a significant amount of memory is required for storing the network's weights and millions of multiplications must be carried out for a single input. Another reason that makes model compression necessary is the low-latency requirement in time-critical scenarios since mobile robots should act immediately in a hazardous situation.

Several different methods for compressing a model are discussed in the literature with Pruning and Knowledge Distillation being the most prevalent ones. Pruning \cite{dong2017learning} is a model compression technique that compresses the original model by removing and eliminating redundant operations, and Knowledge Distillation \cite{hinton2015distilling} is a model compression technique in which the large complex model is replaced by a smaller and less complex one. Moreover, DNN (Deep Neural Network) Splitting is a model compression method in which the computer vision model is divided into two parts, the ``head'' and the ``tail''.
The first part is executed on the local processor of a robot, and the second part, which is more complex than the first one is executed on the edge \cite{matsubara2020head}. Additionally, Neural Filter \cite{matsubara2020neural} is a filter that is embedded in the early layers of the overall object detection model and is a classifier whose output is binary, indicating whether an image is empty or not. Empty images are discarded and are not processed by the model and as a result, fewer images are processed leading to a reduced amount of total calculations completed.

When it comes to Pruning, there are numerous techniques that one could use that follow different criteria. Minimum weight, activation, and mutual information \cite{molchanov2016pruning} are three main criteria that are used in Pruning. In the minimum weight criterion, connections that have the lowest weights in a layer are removed. In the activation criterion, weights are removed based on the assumption that if an activation value (an output feature map) is low, then this feature detector is not significant for the prediction task, and thus, it is removed. In the mutual information criterion though, the decision of whether to keep a connection or not is based on measuring how much information from one connection is already present in another connection of the neural network. The connection is removed if this information is incorporated in other connections.

Knowledge Distillation introduced in \cite{hinton2015distilling}, is a method in which a simpler model, called the ``student'' model, is trained to mimic the output of a more complex one, called the ``teacher'' model. The key assumption is that large ``teacher'' models are often over-parameterized and can be compressed without any significant performance loss. At the end of the training phase, the ``student'' model replaces the ``teacher'' model and as a result, the complexity and the inference time of the computer vision process are reduced.

These two model compression techniques have been investigated in depth in the literature and have strong theoretical and practical foundations, but work on their combinations is limited. The novel part of this work is to investigate their combination and verify if it outperforms the stand-alone versions.

\section{Background on the Applied Methods}
\label{ch:methods}
In this section, the model compression techniques that are applied in the proposed solution are described, beginning with Pruning and followed by Knowledge Distillation.

\subsection{Pruning}
\label{sec:PruningMethod}
Deep neural networks are often over-parameterized and may have redundant operations that do not have an effect on the end result. 
Pruning is a model compression technique that aims to reduce the number of parameters of complex models by identifying and eliminating redundant operations. 
The first step of this technique is to train the neural network, while the second step is to identify and remove unimportant connections. In this step, connections that have low weights are removed, using one of the many different Pruning techniques that are available. The final step of Pruning is the fine-tuning of the obtained model to avoid losing a significant amount of accuracy.

\begin{figure}[!ht]
 \centering
  \includegraphics[width=87mm]{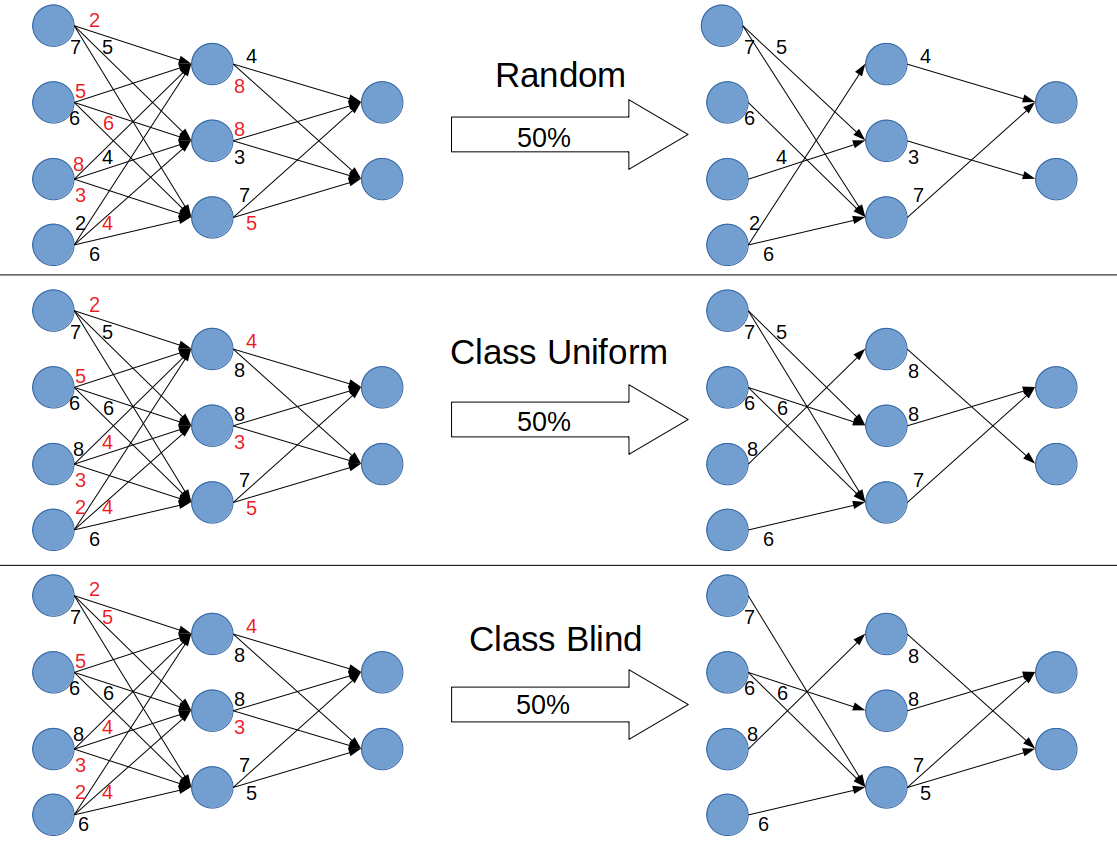}
  \caption{Different Pruning techniques result in different end models, even when using the same initial neural network. Top figure shows random Pruning in which 50\% connections are removed randomly. Middle figure shows class uniform where 50\% lowest weights are discarded. Bottom figure shows class blind which 50\% lowest weights of each layer are eliminated.}
  \label{fig:Pruning-screenshot}
\end{figure}

Among existing Pruning techniques, this work employs random, class-uniform, and class-blind \cite{marchisio2018prunet} Pruning. In random Pruning, a percentage $x$ is selected by the user, and regardless of their layer and weight value, $x \%$ connections are randomly removed from the whole network. In class-uniform, Pruning is distributed equally among hidden layers, which is achieved by selecting a percentage $x$ and Pruning out the $x \%$ connections with the lowest weights per layer. Whereas in class-blind, given a percentage $x$, connections with the lowest $x \%$ are removed from the network regardless of the layer they are in. Studies showed a better performance of class-blind compared to the previous two techniques \cite{marchisio2018prunet}. Each Pruning technique results in a different network as can be seen in \autoref{fig:Pruning-screenshot}. Connections that are going to be pruned are marked with red labels.

\subsection{Knowledge Distillation}
\label{sec:KnowlDistilMethod}
\begin{figure}[ht!]
 \centering
  \includegraphics[width=80mm]{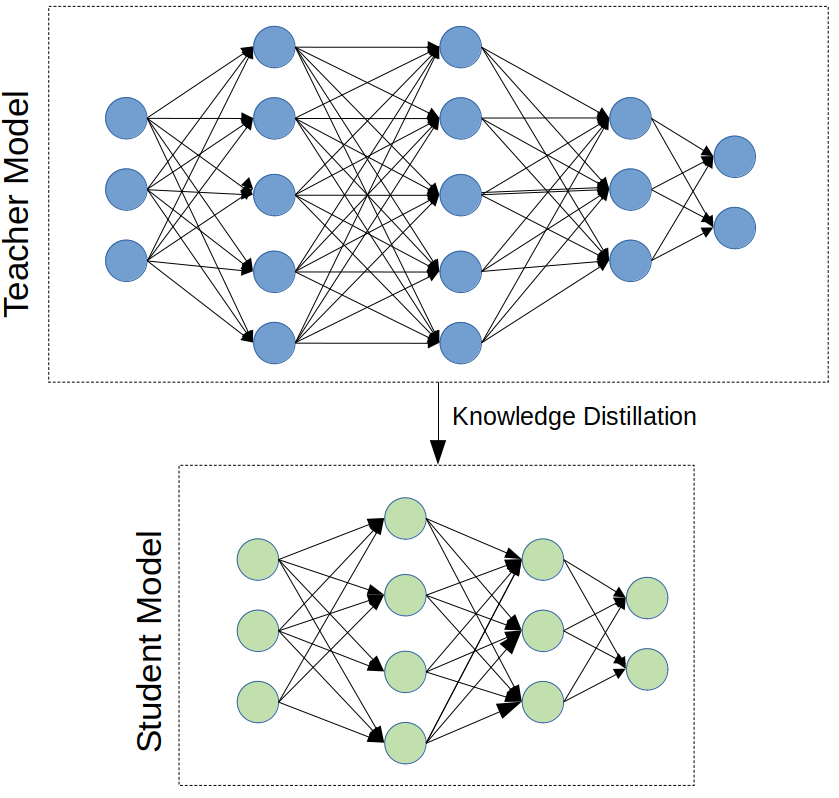}
  \caption{Knowledge Distillation example that begins from a large complex teacher model and results in a smaller, less complex student model.}
  \label{fig:DNN-KD-screenshot}
\end{figure}

The goal of Knowledge Distillation is to use a less complex model, so-called the student model, to replicate the output represented by a more complex model, so-called the teacher model. This method is based on the assumption that teacher models are often over-parameterized, and can be compressed without causing significant performance loss. By using a simpler model that produces the same output, the total inference time is reduced. An example of Knowledge Distillation compression can be seen in \autoref{fig:DNN-KD-screenshot}. 
Mimicking the output of the original model is the key to a successful Knowledge Distillation, and when the student model replicates the output of a teacher model, the knowledge transferred from it can achieve higher performance than the original one \cite{cho2019efficacy}. This replication is achieved by training the student model using the output of the teacher model instead of using the labels provided by a dataset. For performing the Knowledge Distillation technique, both models, the teacher and the student must be selected by the user. The selection of these models can be challenging as it can be difficult to find an appropriate student model. Therefore, a trial-and-error process may be necessary to ensure satisfactory results.

\section{Experimental Setup}
\label{ch:experiments}
In total, three sets of experiments with the same goal are implemented. This goal is to reduce the total size of the selected models without losing any significant amount of accuracy. The first set of experiment that is implemented is Pruning, with the second one being Knowledge Distillation. The third and last set of experiment is a combination of the first two experiments, which consists of applying Knowledge Distillation on a selected model and then using Pruning on the resulting model. The experiments are reproduced for two different tasks, Object Detection and Semantic Segmentation.

\subsection{Datasets}

For the Object Detection task, where the output of the computer vision models are
bounding boxes of the detected objects, the dataset that is used is CIFAR10 \cite{krizhevsky2009learning}. 

For the Semantic Segmentation part of the experiments, where the output of the computer vision models are segmentation masks of the detected objects, the datasets that are used are COCO \cite{lin2014microsoft} and Pascal-VOC \cite{pascal-voc-2012}. 

These datasets were chosen as they are easily accessible through common libraries, such as torchvision \cite{paszke2017automatic}.

\subsection{Models}

For the purposes of this work, five models are compressed using two model compression techniques. Three of the models are used for the object detection task, which are LeNet \cite{yu2015handwritten} and two models from the ResNet family \cite{DBLP:journals/corr/HeZRS15}, which are ResNet-$18$ and ResNet-$101$. The remaining two models are used for the semantic segmentation task and these models are DeepLabV$3$ MobileNetV$3$ \cite{chen2019rethinking} and one from the ResNet family, FCN ResNet-$101$. These models were chosen due to their number of parameters and considerable differences between them. For this work, any model with a similar number of parameters might be used, but these five were chosen due to their easy accessibility and widespread use.

The total number of parameters for each of these models is listed in \autoref{tab:NumOfParamsInitial}.

\begin{table}[H]
\caption{Number of parameters of the models used.}
\begin{subtable}{\linewidth}
\centering
\begin{tabular}{|m{2.5cm}|m{2.5cm}|m{2.5cm}|}
\hline
\multicolumn{3}{|c|}{\textbf{Object Detection}} \rule{0pt}{0.5cm}\\ \hline
\textbf{LeNet} & \textbf{ResNet-18} & \textbf{ResNet-101}\rule{0pt}{0.3cm} \\ \hline
$0.357 \times 10^6$ & $11.2\times 10^6$ & $44.5\times 10^6$  \rule{0pt}{0.3cm}\\ \hline 
\end{tabular}

\vspace{0.5cm}

\begin{tabular}{|m{3.501cm}|m{4.49cm}|}
\hline
\multicolumn{2}{|c|}{\textbf{Semantic Segmentation}}  \rule{0pt}{0.5cm}\\ \hline
 \textbf{FCN ResNet-101} &  \textbf{DeepLab3} \rule{0pt}{0.3cm}\\ \hline
$54.3\times 10^6$ & $11.1\times 10^6$  \rule{0pt}{0.3cm}\\ \hline
\end{tabular}
\end{subtable}
\label{tab:NumOfParamsInitial}
\end{table}

\subsection{Performance Metric}
The metric that is considered for determining which model performs best is a trade-off between the drop in the accuracy of the compressed model and the percentage that the network was compressed. The goal is to maximize the compression percentage while minimizing the accuracy drop of the compressed model (for the Object Detection, top-$1$ accuracy is considered, and for the Semantic Segmentation task, Global Pixel Accuracy is considered \cite{8500497}).

\section{Results and Analysis}
\label{ch:resultsAndAnalysis}
In this section, the results of the three experiments are presented and analyzed: Pruning, Knowledge Distillation and the combination of these two techniques.

\subsection{Pruning}
In this experiment, three different Pruning techniques are applied in the different models presented in \autoref{tab:NumOfParamsInitial}, aiming to identify the best performing Pruning technique for both Object Detection and Semantic Segmentation tasks. The three Pruning techniques that are used are random, class-uniform and class-blind Pruning.

\begin{figure}[ht!]
\centering
\begin{subfigure}[!htbp]{0.33\linewidth}
    \centering
    \includegraphics[width=\linewidth, trim = 4.5cm 8cm 4cm 8cm, clip]{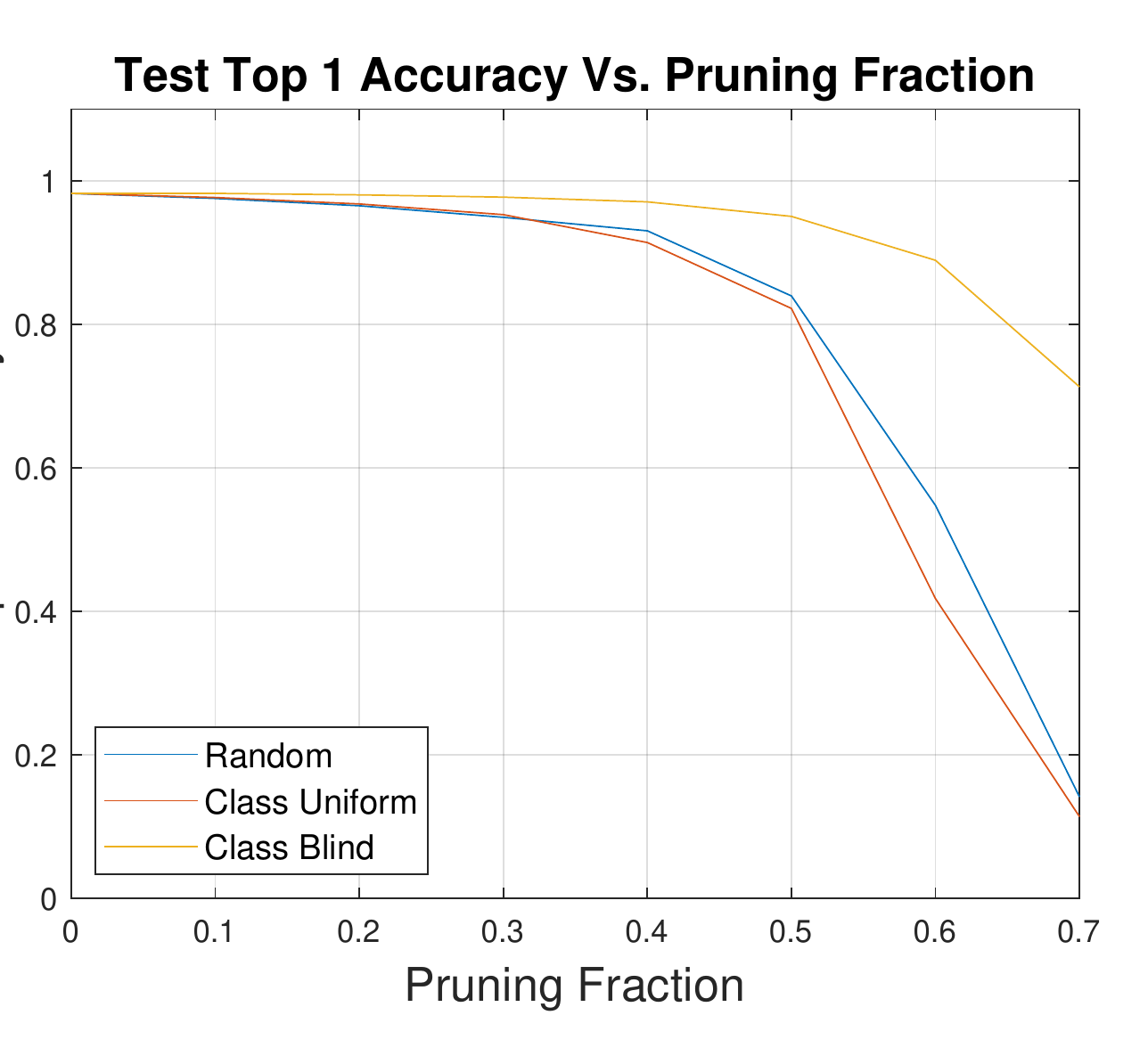}
    \caption{Results for LeNet.}
    \label{fig:PruningLeNet}
\end{subfigure}%
\begin{subfigure}[!htbp]{0.33\linewidth}
    \centering
    \includegraphics[width=\linewidth, trim = 4.5cm 8cm 4cm 8cm, clip]{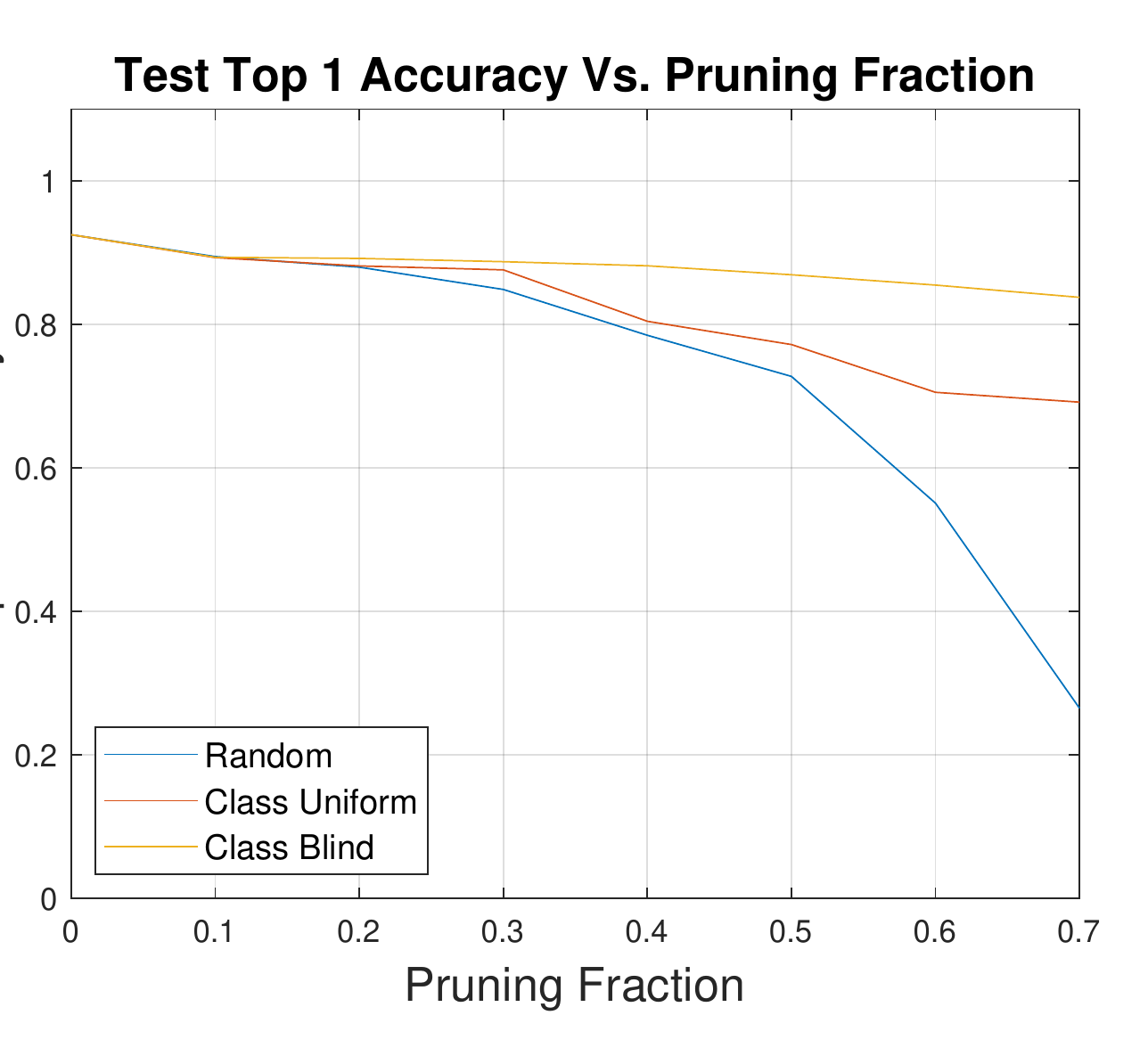}
    \caption{Results for ResNet-$18$.}   
    \label{fig:PruningResNet18}
\end{subfigure}%
\begin{subfigure}[!htbp]{0.33\linewidth}
    \centering
    \includegraphics[width=\linewidth, trim = 4.5cm 8cm 4cm 8cm, clip]{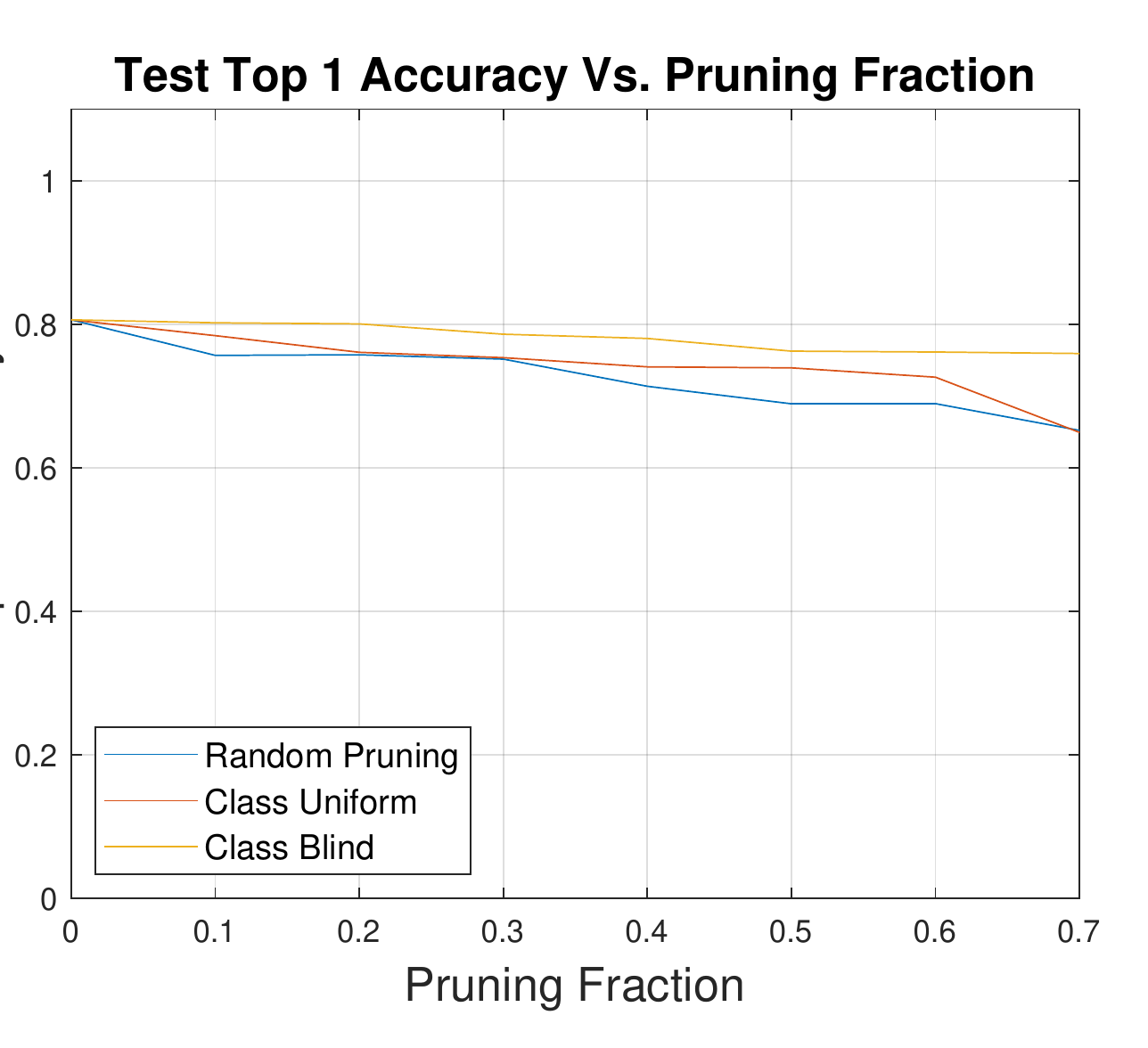}
    \caption{Results for ResNet-$101$.}   
    \label{fig:PruningResNet101}
\end{subfigure}
\caption{Top-1 Accuracy for Object Detection models for different Pruning strategies and percentages.}
\label{fig:PruningResults}
\end{figure}

\begin{figure*}[ht!]
\centering
\begin{subfigure}[!htbp]{0.33\linewidth}
    \centering
    \includegraphics[width=\linewidth, trim = 4.5cm 8cm 4cm 8cm, clip]{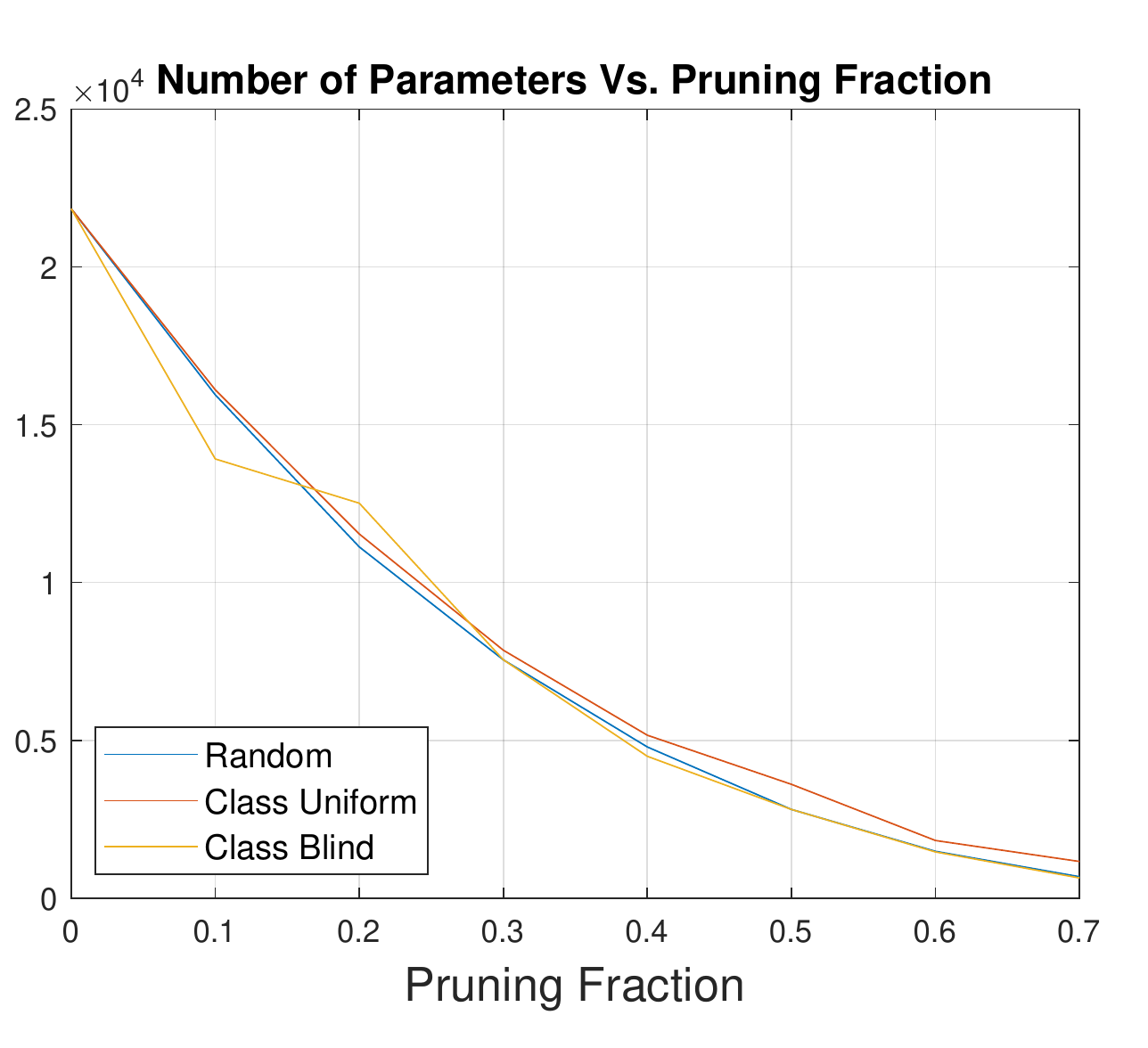}
    \caption{Results for LeNet.}
    \label{fig:PruningNumLeNet}
\end{subfigure}%
\begin{subfigure}[!htbp]{0.33\linewidth}
    \centering
    \includegraphics[width=\linewidth, trim = 4.5cm 8cm 4cm 8cm, clip]{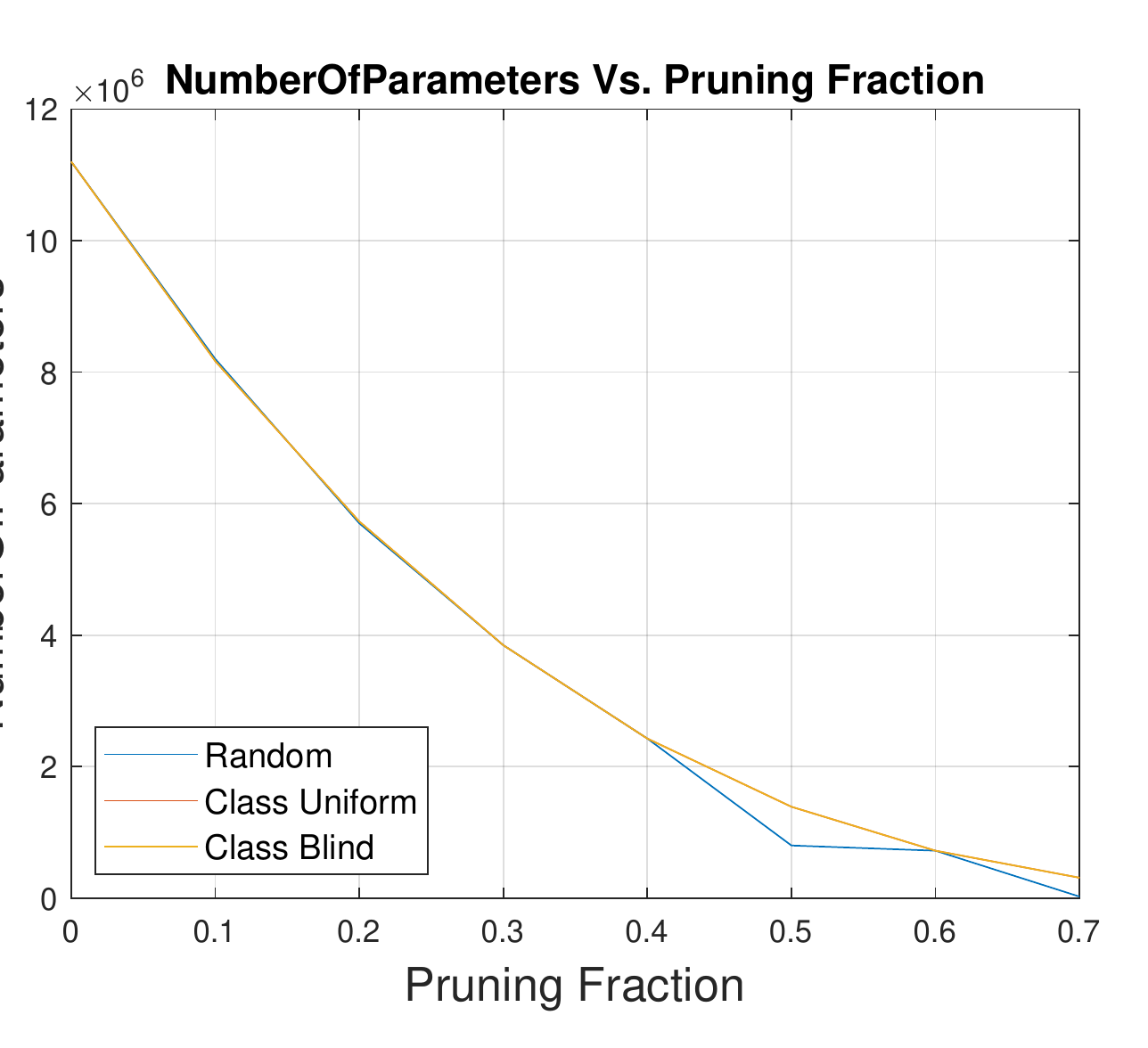}
    \caption{Results for ResNet-$18$.}   
    \label{fig:PruningNumResNet18}
\end{subfigure}%
\begin{subfigure}[!htbp]{0.33\linewidth}
    \centering
    \includegraphics[width=\linewidth, trim = 4.5cm 8cm 4cm 8cm, clip]{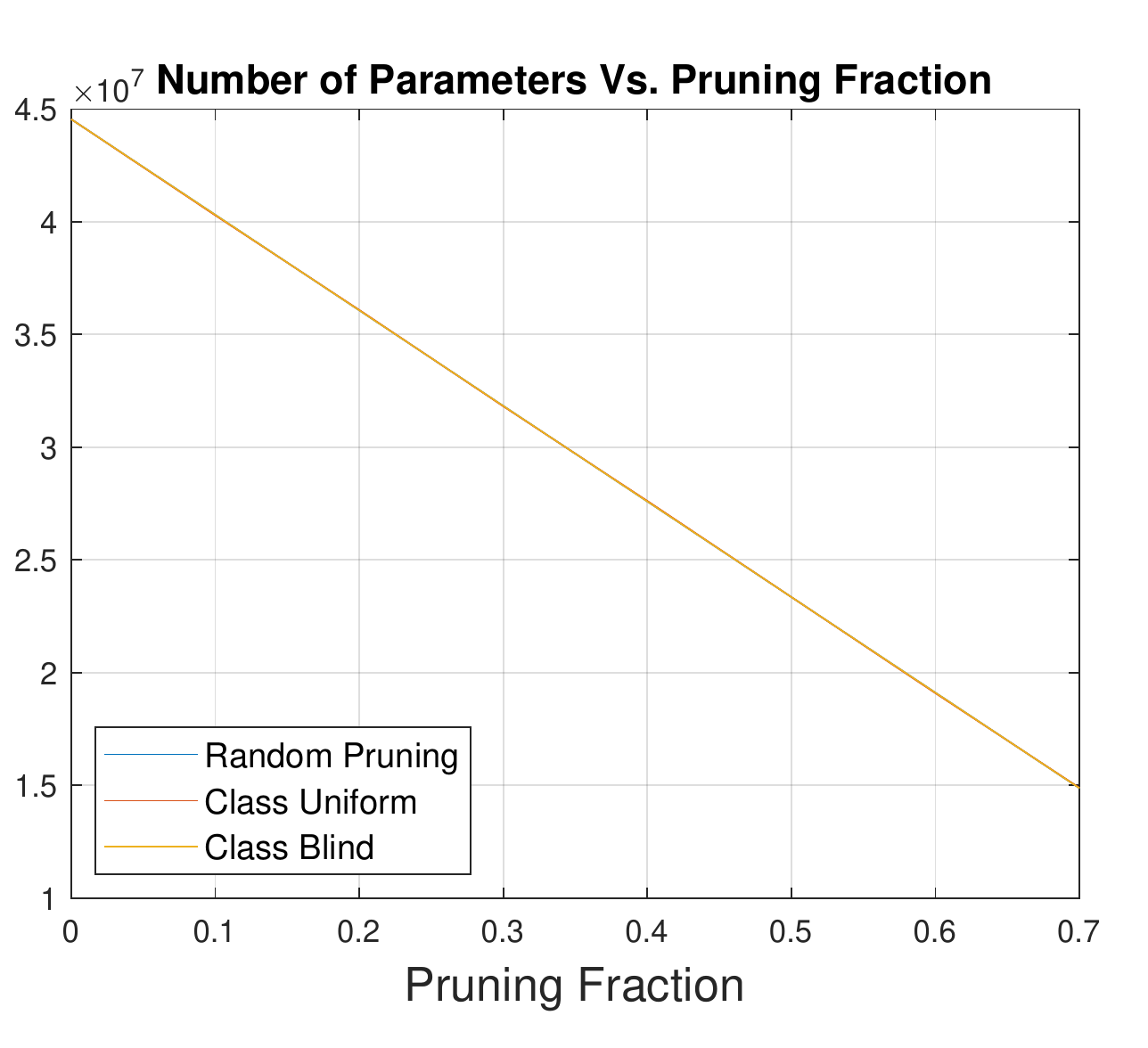}
    \caption{Results for ResNet-$101$.}   
    \label{fig:PruningNumResNet101}
\end{subfigure}
\caption{Number of Parameters for Object Detection models for different Pruning strategies and percentages.}
\label{fig:PruningResultsNumOfParams}
\end{figure*}

The Pruning results for the object detection models can be seen in \autoref{fig:PruningResults} and \autoref{fig:PruningResultsNumOfParams}, which present the top-$1$ validation accuracy and the number of parameter curves, respectively. \autoref{fig:PruningLeNet} and \autoref{fig:PruningNumLeNet} present the results for the LeNet model, while \autoref{fig:PruningResNet18} and \autoref{fig:PruningNumResNet18} present the results for ResNet-$18$ model, and \autoref{fig:PruningResNet101} and \autoref{fig:PruningNumResNet101} present the results for the ResNet-$101$ model. These results show that in all three models and for most of the different Pruning percentages, the technique that performed best was the class-blind technique, as it achieved higher accuracy than the other two techniques. Furthermore, the class-blind technique was more consistent than the other two techniques, as the accuracy of the models did not have a large drop, even for Pruning percentages above $40 \%$. Thus, it can be concluded that the class-blind technique is the best performing Pruning technique for the object detection task and will be used for the upcoming experiments.

As for the number of parameters, a linear behavior among the different Pruning percentages was expected, as well as the same number of parameters for all three Pruning techniques since the same Pruning percentage was applied. However, as it can be observed from \autoref{fig:PruningResultsNumOfParams}, this is not the case. This non-linear behavior takes place as in some cases, all the input connections of a neuron are pruned, and as a result, all the output connections are pruned as well since there is no input. This is the main reason for the difference in the number of parameters between different Pruning techniques and for the non-linear behavior for different Pruning percentages.

\begin{figure*}[ht!]
\centering
\begin{subfigure}[!htbp]{0.4\linewidth}
    \centering
    \includegraphics[width=\linewidth, trim = 3cm 8cm 3cm 8cm, clip]{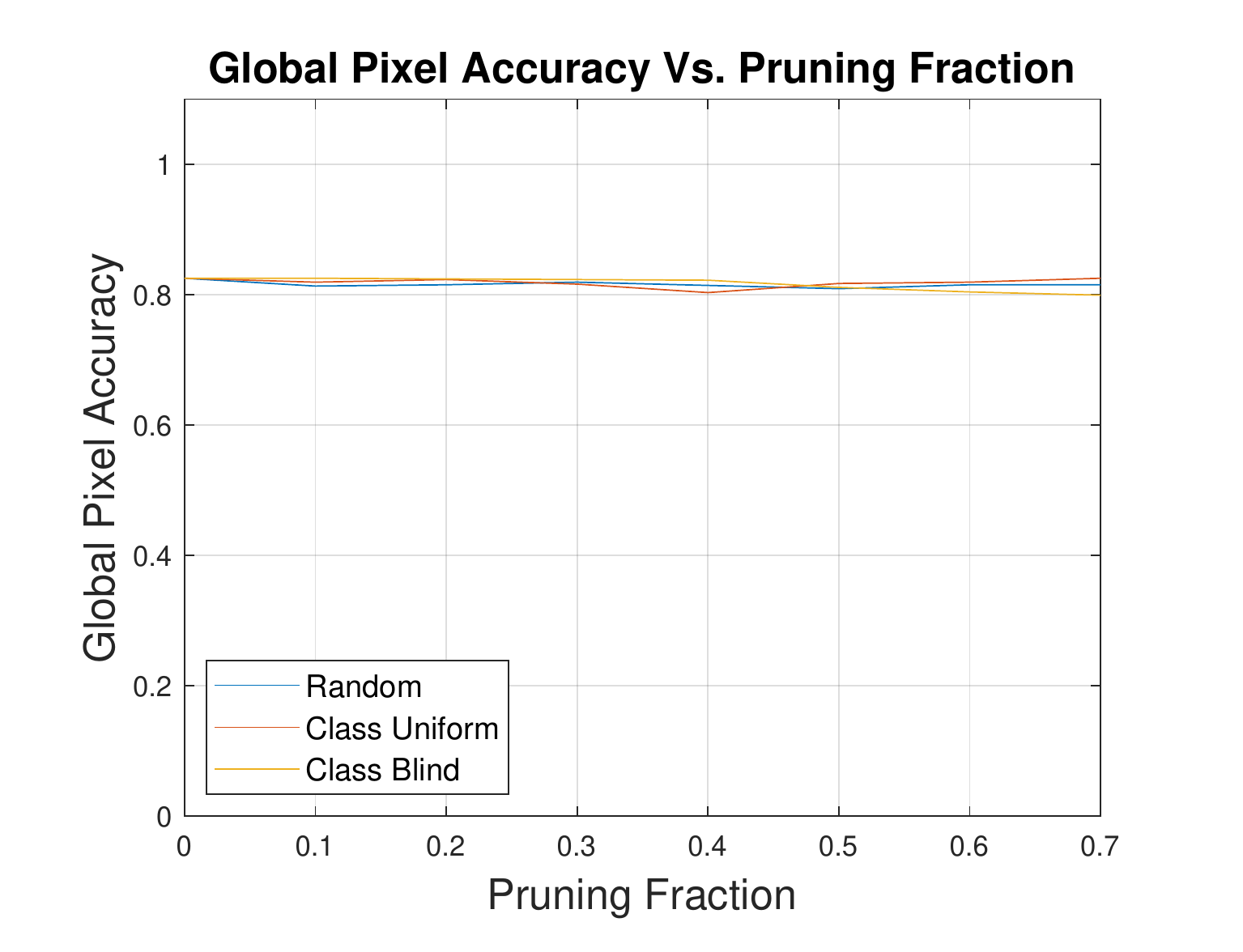}
    \caption{Results for COCO Dataset.}
    \label{fig:COCOPruning}
\end{subfigure}%
\begin{subfigure}[!htbp]{0.4\linewidth}
    \centering
    \includegraphics[width=\linewidth, trim = 3cm 8cm 3cm 8cm, clip]{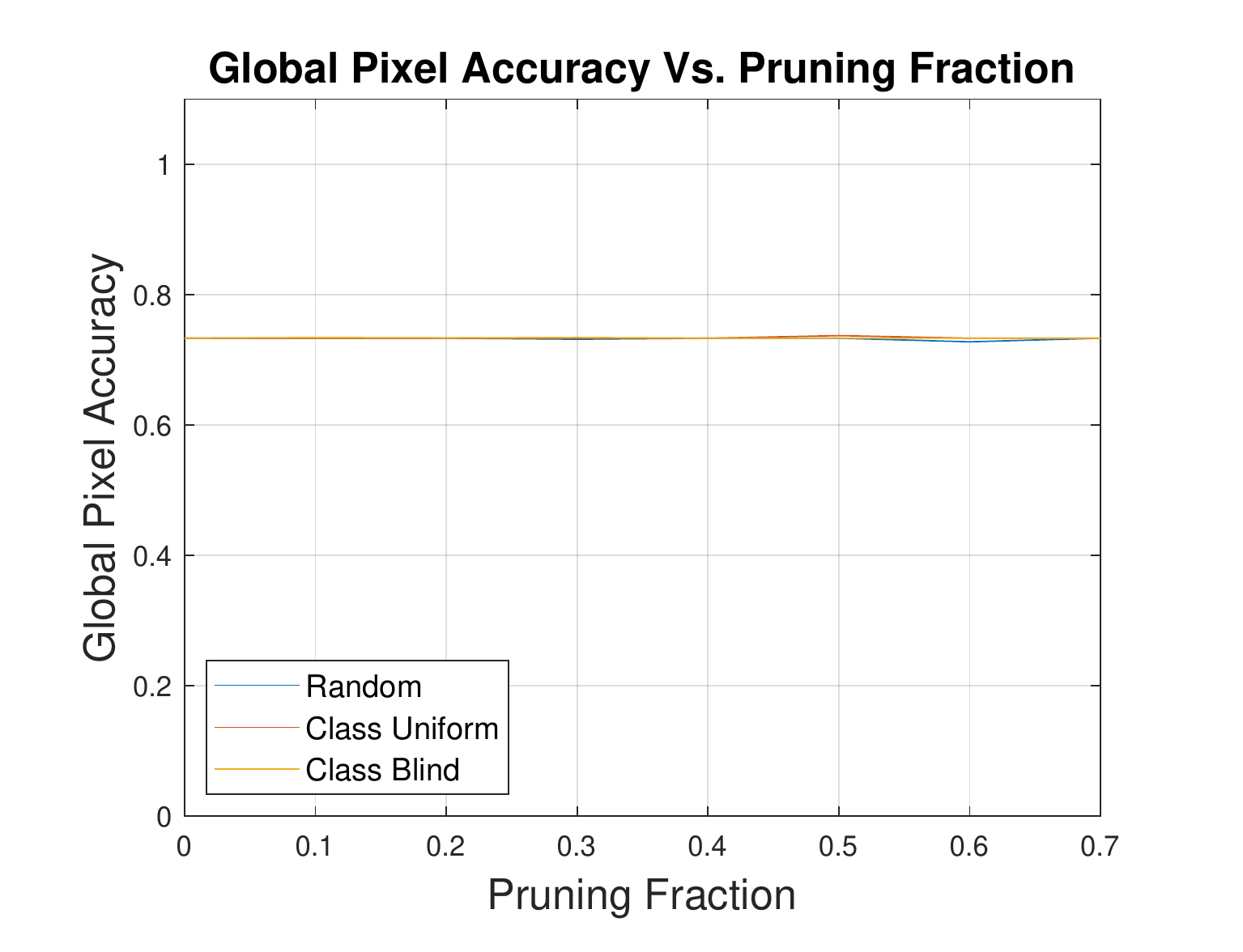}
    \caption{Results for VOC Dataset.}   
    \label{fig:VOCPruning}
\end{subfigure}
\caption{Global Pixel Accuracy for Semantic Segmentation for different Pruning strategies and percentages.}
\label{fig:SegmentationPruning}
\end{figure*}

The results for the semantic segmentation models can be seen in \autoref{fig:SegmentationPruning}, which presents the Global Pixel Accuracy of the MobileNet model for the different Pruning strategies and percentages. \autoref{fig:COCOPruning} and \autoref{fig:VOCPruning} present the results for the two different datasets used in this experiment. The semantic segmentation results show that there are no significant differences in the accuracy drop between the three Pruning techniques, even for Pruning percentages above $40 \%$. Thus, it can be concluded that any of these techniques could be used for the upcoming experiments, but the decision was to use the class-blind technique in order to stay consistent between object detection and semantic segmentation tasks.

\subsection{Knowledge Distillation}
In the Knowledge Distillation experiment, different combinations of models were tested to evaluate the accuracy drop between the teacher and the student models. In total, four different combinations were evaluated, three for the object detection task and one for the semantic segmentation task. At the end of this experiment, the best combinations out of them are selected and used for the upcoming experiment.

For the object detection task, the first evaluated combination had ResNet-$18$ as the teacher model and LeNet as the student model, while the second one used ResNet-$101$ as the teacher model and ResNet-$18$ as the student model. The third combination for this task used ResNet-$101$ as the teacher model and LeNet as the student model. For the semantic segmentation task, one combination was tested out and that was using FCN ResNet-$101$ as the teacher model and MobileNet-V$3$ as the student model, both for COCO and VOC datasets.

In each combination, the accuracy and the number of parameters of the student model were obtained. These values along with the accuracy of the teacher model are presented in \autoref{tab:KnowlDistillResults} and \autoref{tab:SemanticSegmKnowlDistil} for object detection and semantic segmentation, respectively. These results show an accuracy higher than $80 \%$ for all combinations while significantly reducing the number of parameters. We see that in three combinations (ResNet-101 to ResNet-18, ResNet-101 to LeNet and FCN ResNet-101 to MobileNet-V3), the student model achieved higher accuracy than the teacher model. This behavior is in line with the literature \cite{cho2019efficacy} and can happen when the student model has a more suitable DNN architecture for a given task than the teacher model.

\begin{center}
\begin{table}[H]
\centering
\begin{tabular}{|m{4.5cm}|m{3cm}|m{3cm}|m{3cm}|}
\hline
 & \textbf{ResNet-18 to LeNet  } & \textbf{ResNet-101 to ResNet-18} & \textbf{ResNet-101  to LeNet } \rule{0pt}{0.4cm} \\ \hline
\textbf{Accuracy of the teacher model(\%)}  & $92.48$ & $80.62$ & $80.62$\\ \hline
\textbf{Accuracy of the student model(\%)}  & $85.10$  & $86.81$ & $81.47$\\ \hline
\textbf{Number of Parameters}               & $11.2 \times 10^6$ to $0.357 \times 10^6$ & $44.5 \times 10^6$ to $11.2 \times 10^6$ & $44.5 \times 10^6$ to $0.357\times 10^6$ \rule{0pt}{0.4cm}\\ \hline
\end{tabular}
\caption{Results of the validation accuracy for different combinations of teacher and student models for the object detection task.}
\label{tab:KnowlDistillResults}
\end{table}
\end{center}

\begin{center}
\begin{table}[H]
\centering
\begin{tabular}{|m{4.5cm}|m{3cm}|m{3cm}|}
\hline
 & \textbf{VOC Dataset} & \textbf{COCO Dataset} \\ \hline
\textbf{Accuracy of the teacher model(\%)}  & $82.4$ & $82.4$\\ \hline
\textbf{Accuracy of the student model(\%)}  & $88.06$  & $90.68$\\ \hline
\textbf{Number of Parameters}               & $54.3 \times 10^6$ to $11.1\times 10^6$ & $54.3\times 10^6$ to $11.1\times 10^6$\rule{0pt}{0.35cm}\\ \hline
\end{tabular}
\caption{Results of the validation accuracy for combination of teacher model being FCN ResNet-$101$ and the student model being MobileNet-V$3$ for the semantic segmentation task.}
\label{tab:SemanticSegmKnowlDistil}
\end{table}
\end{center}
\vspace{0.1cm}

Finally, from the Knowledge Distillation experiment, all four combinations achieved accuracy higher than $80 \%$, and the total number of parameters is reduced more than $75 \%$. Nevertheless, the combinations that are going to be used for the upcoming experiment are ResNet-$18$ to LeNet and ResNet-$101$ to ResNet-$18$ for the object detection task, and FCN ResNet-$101$ to MobileNet-V$3$ for the semantic segmentation task. These combinations were selected as they achieve total accuracy higher than $85 \%$ and they have a sufficient number of parameters left for investigating the effect of Pruning when applied to the student model.

\subsection{Combining Knowledge Distillation and Pruning}
\begin{figure*}[ht!]
\centering
\begin{subfigure}[!htbp]{0.32\linewidth}
    \centering
    \includegraphics[width=\linewidth, trim = 4.5cm 8cm 4cm 8cm, clip]{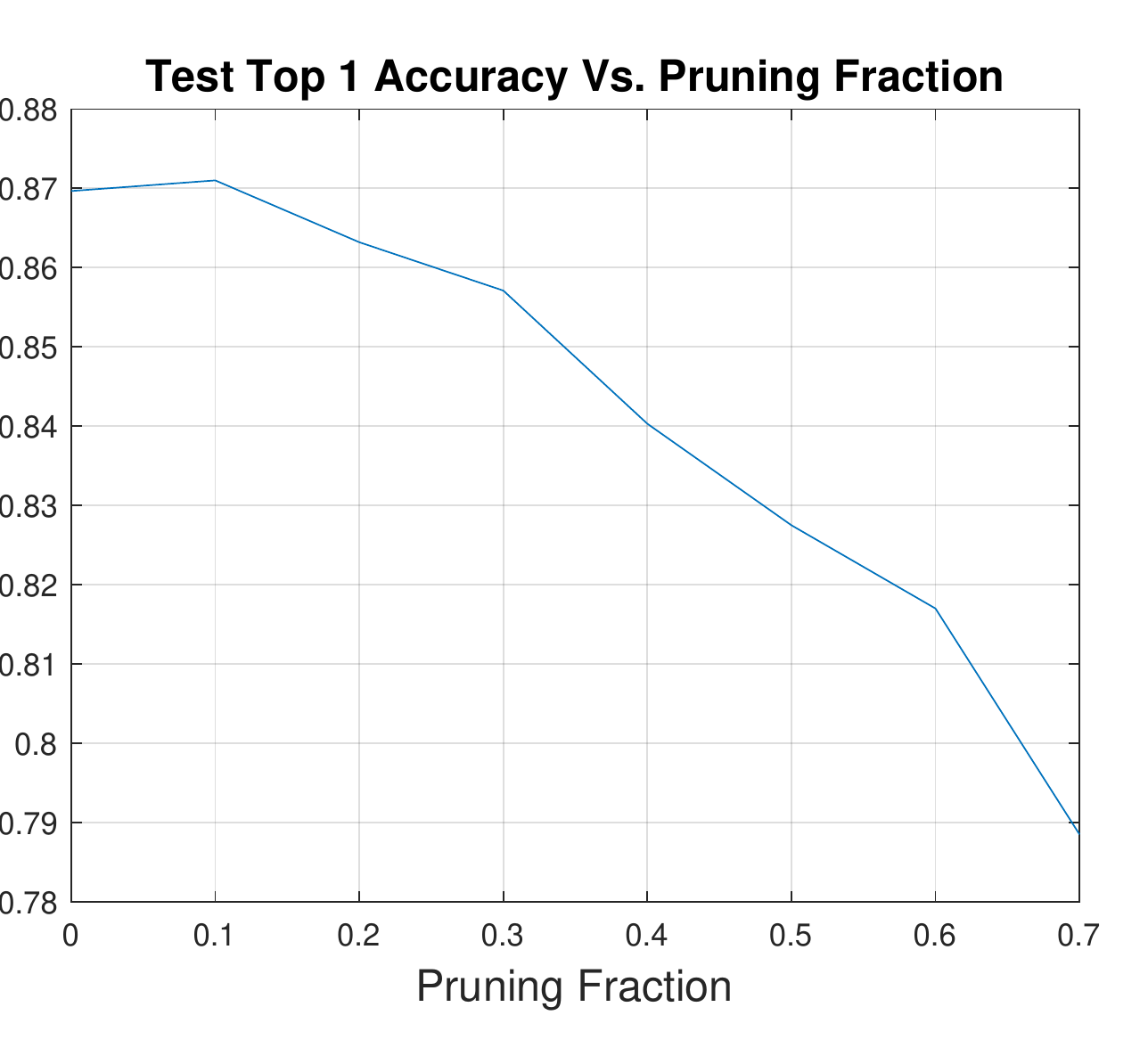}
    \caption{Results for ResNet-$18$ model to LeNet.}
    \label{fig:ComboResnet18toLeNet}
\end{subfigure}%
\hfill
\begin{subfigure}[!htbp]{0.32\linewidth}
    \centering
    \includegraphics[width=\linewidth, trim = 4.5cm 8cm 4cm 8cm, clip]{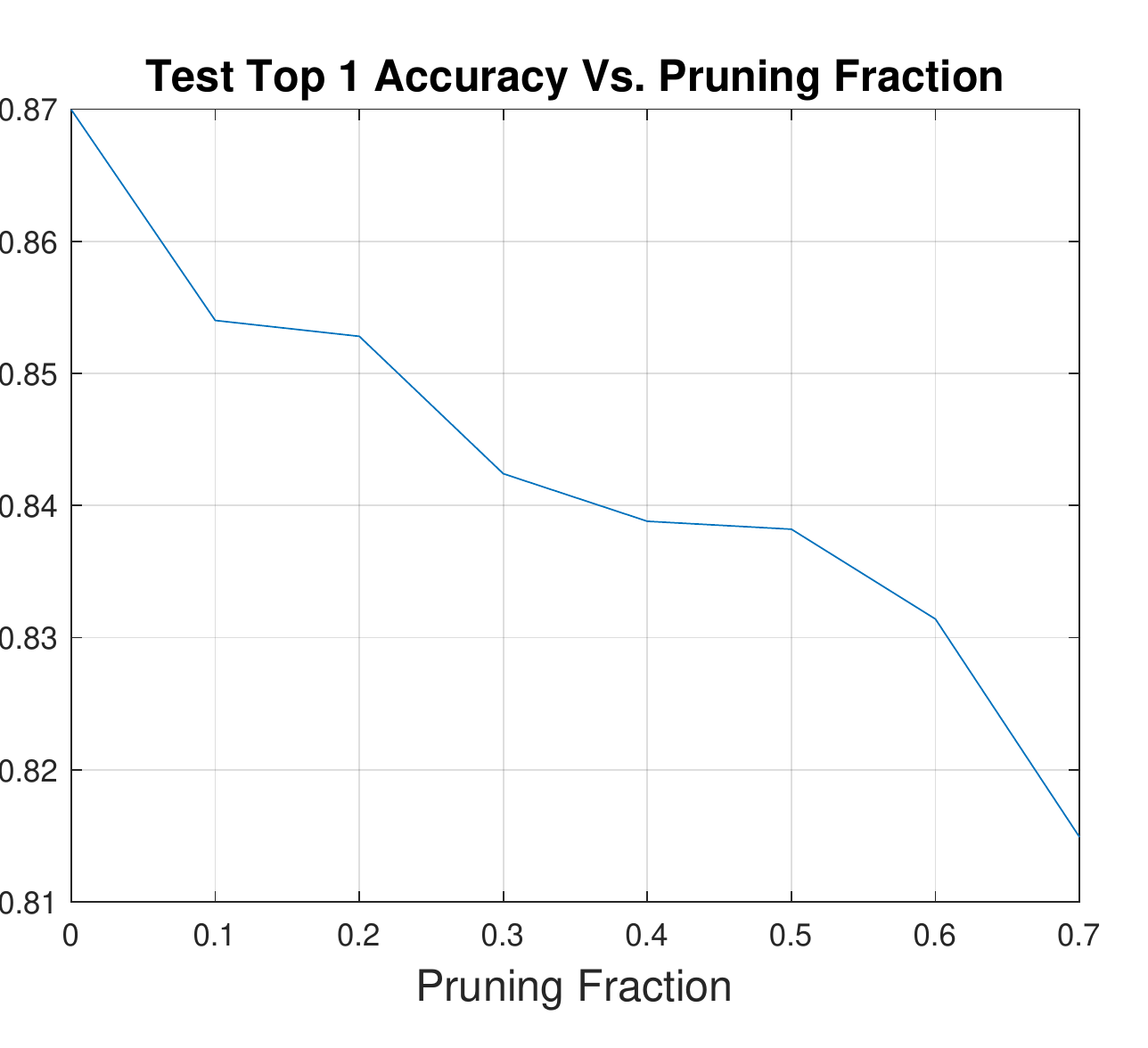}
    \caption{Results for ResNet-$101$ to ResNet-$18$.}   
    \label{fig:ComboResnet101toResNet18}
\end{subfigure}%
\hfill
\begin{subfigure}[!htbp]{0.32\linewidth}
    \centering
    \includegraphics[width=\linewidth, trim = 4.5cm 8cm 4cm 8cm, clip]{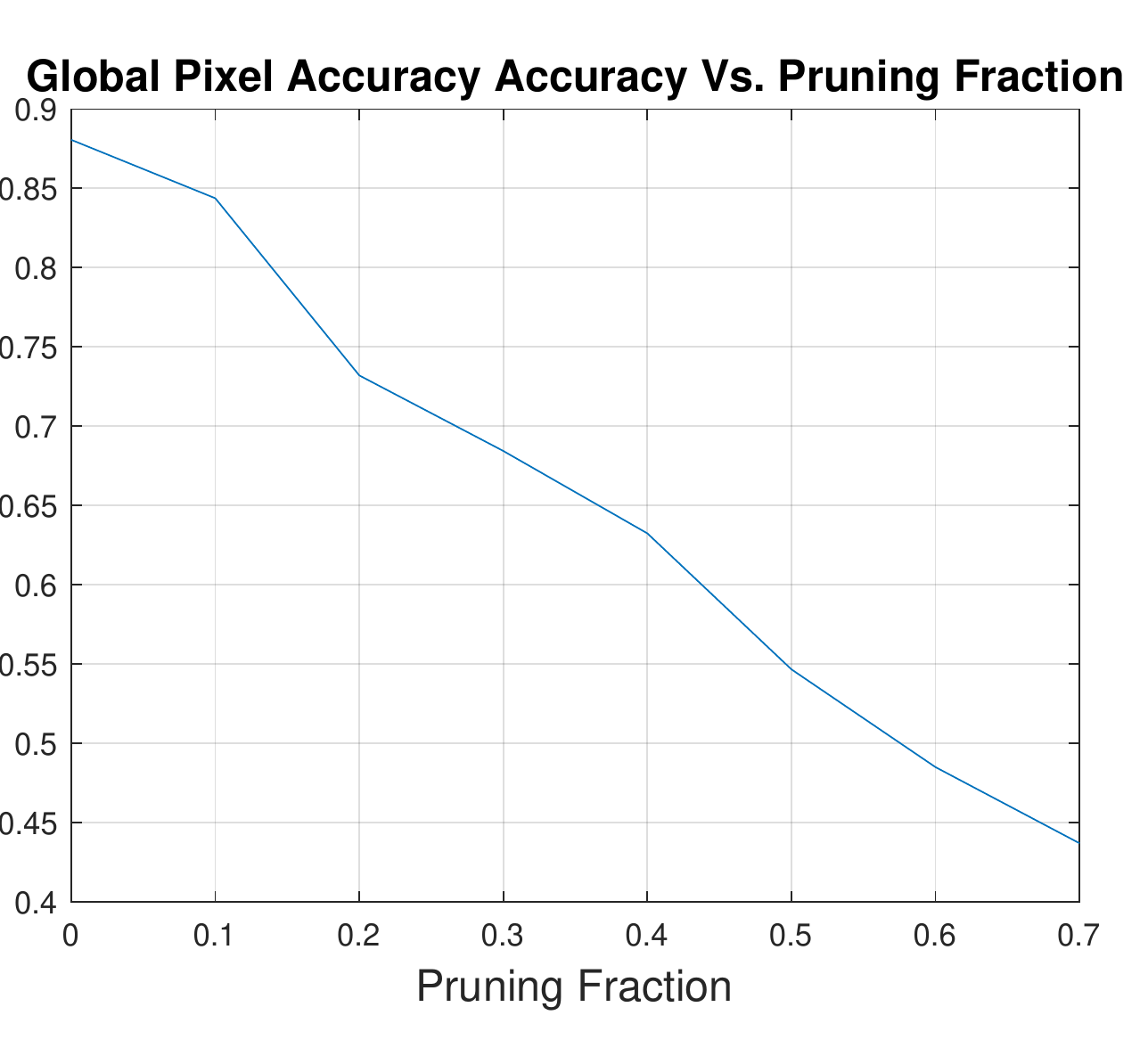}
    \caption{Results for FCN ResNet-$101$ to MobileNet-V$3$.}   
    \label{fig:ComboSemantic}
\end{subfigure}
\caption{Results for the Pruning part for the three resulted student models.}
\label{fig:ComboPruning}
\end{figure*}

In this experiment, which is the main contribution of this work, Knowledge Distillation is first applied on a large teacher model, and then Pruning is applied on the resulting student model. Based on the results of the two previous experiments, three student models from Knowledge Distillation were selected and then applied class-blind Pruning. The resulting plots for the pruned student models with the accuracy values against the Pruning fraction are shown in \autoref{fig:ComboPruning}.

From the results of the object detection task, which can be seen in the first two rows of \autoref{tab:KnowlDistillResults} and from figures \autoref{fig:ComboResnet18toLeNet} and \autoref{fig:ComboResnet101toResNet18}, it can be observed that the results of both combinations achieved total accuracy higher than $80 \%$ for a compression percentage of more than $90 \%$. An even higher compression percentage can be achieved by applying a bigger Pruning fraction, but it comes with the cost of total accuracy being less than $80 \%$. The best-performing combination is considered to be the second one, which is using ResNet-$101$ as the teacher model and ResNet-$18$ as the student model. The reason is that it has a lower accuracy drop than the other combination, with approximately the same ratio in the number of parameters between the student and the teacher model.

The results of the semantic segmentation task can be seen in \autoref{tab:SemanticSegmKnowlDistil} and from figure \autoref{fig:ComboSemantic}. From these results, it can be observed that for a Pruning percentage larger than $20 \%$, there is a significant drop in the accuracy of the resulting model. Thus, if the total accuracy of the model needs to be higher than $80 \%$, then the initial model cannot be compressed more than $85 \%$, which is achieved by applying $10 \%$ Pruning percentage when Pruning the student model.

\section{Discussion}
\label{ch:discussion}
According to the findings of this study, the model compression techniques evaluated were able to retain high levels of accuracy (above $80 \%$) even when the compression percentage was greater than $90 \%$. As a result, it can be argued that the size of a computer vision model, or a DNN in general, can be reduced without sacrificing a significant amount of accuracy, confirming the hypothesis that large computer vision models are frequently over-parameterized. Furthermore, Pruning and Knowledge Distillation model compression techniques were proven to be effective when applied individually to computer vision algorithms and DNNs, and they had even better results when they were combined.

Despite these observations, the proposed solution have limitations. Currently it is not possible to generalize our findings for any computer vision model as each model has its own particularities. Therefore, a manual trial-and-error approach should be performed for each case. Based on the results of this work, a simple recommendation for finding a proper student model in a Knowledge Distillation experiment is to avoid models with a ratio of less than $20 \%$ between the number of parameters of the student and the teacher model, implying that the number of parameters should not be reduced by more than $80 \%$.

\section{Conclusion and Future Works}
\label{ch:conclusionsAndFutureWork}
The purpose of this work was to reduce the complexity of computer vision models while maintaining high levels of accuracy. The results of this attempt were successful in reducing the complexity and the size of these models while maintaining the accuracy at high levels. In this work, we have shown that the size of a computer vision model can be reduced by $90 \%$ while still achieving accuracy higher than $80 \%$. This leads to fewer computations to obtain the result, which means that lower inference time is achieved, and a reduction in computer resources as well. 
Consequently, real-time processing may be achieved with a compressed version of a model when computations are done locally on the mobile robot's processor. In cases where high precision is required, the original model might be preferred over a compressed version of it.

Due to the scope of this problem, there is room for improvement and future work. A possible future work will be to employ more model compression techniques, such as DNN splitting and neural filtering, and perform more experiments, or even test more combinations for the two techniques that were used in this work. One more possible future work is to evaluate the best-performing model in a real-world scenario and determine if the compressed version of the model produces the desired output without being connected to edge resources. Finally, one last possible future work is to compare a compressed model against an already smaller model, like YOLO, to determine if compressing a model is better than using a smaller one from the beginning.

\bibliographystyle{eptcs}
\bibliography{generic}

\begin{thebibliography}{10}
\providecommand{\bibitemdeclare}[2]{}
\providecommand{\surnamestart}{}
\providecommand{\surnameend}{}
\providecommand{\urlprefix}{Available at }
\providecommand{\url}[1]{\texttt{#1}}
\providecommand{\href}[2]{\texttt{#2}}
\providecommand{\urlalt}[2]{\href{#1}{#2}}
\providecommand{\doi}[1]{doi:\urlalt{https://doi.org/#1}{#1}}
\providecommand{\eprint}[1]{arXiv:\urlalt{https://arxiv.org/abs/#1}{#1}}
\providecommand{\bibinfo}[2]{#2}

\bibitemdeclare{article}{chen2019rethinking}
\bibitem{chen2019rethinking}
\bibinfo{author}{Liang{-}Chieh \surnamestart Chen\surnameend},
  \bibinfo{author}{George \surnamestart Papandreou\surnameend},
  \bibinfo{author}{Florian \surnamestart Schroff\surnameend} \&
  \bibinfo{author}{Hartwig \surnamestart Adam\surnameend}
  (\bibinfo{year}{2017}): \emph{\bibinfo{title}{Rethinking Atrous Convolution
  for Semantic Image Segmentation}}.
\newblock {\slshape \bibinfo{journal}{CoRR}} \bibinfo{volume}{abs/1706.05587},
  \doi{10.48550/arXiv.1706.05587}.
\newblock \eprint{1706.05587}.

\bibitemdeclare{inproceedings}{cho2019efficacy}
\bibitem{cho2019efficacy}
\bibinfo{author}{Jang~Hyun \surnamestart Cho\surnameend} \&
  \bibinfo{author}{Bharath \surnamestart Hariharan\surnameend}
  (\bibinfo{year}{2019}): \emph{\bibinfo{title}{On the Efficacy of Knowledge
  Distillation}}.
\newblock In: {\slshape \bibinfo{booktitle}{2019 IEEE/CVF International
  Conference on Computer Vision (ICCV)}}, pp. \bibinfo{pages}{4793--4801},
  \doi{10.1109/ICCV.2019.00489}.

\bibitemdeclare{article}{dong2017learning}
\bibitem{dong2017learning}
\bibinfo{author}{Xin \surnamestart Dong\surnameend}, \bibinfo{author}{Shangyu
  \surnamestart Chen\surnameend} \& \bibinfo{author}{Sinno~Jialin \surnamestart
  Pan\surnameend} (\bibinfo{year}{2017}): \emph{\bibinfo{title}{Learning to
  Prune Deep Neural Networks via Layer-wise Optimal Brain Surgeon}}.
\newblock {\slshape \bibinfo{journal}{CoRR}} \bibinfo{volume}{abs/1705.07565},
  \doi{10.48550/arXiv.1705.07565}.
\newblock \eprint{1705.07565}.

\bibitemdeclare{misc}{pascal-voc-2012}
\bibitem{pascal-voc-2012}
\bibinfo{author}{M.~\surnamestart Everingham\surnameend},
  \bibinfo{author}{L.~\surnamestart Van~Gool\surnameend},
  \bibinfo{author}{C.~K.~I. \surnamestart Williams\surnameend},
  \bibinfo{author}{J.~\surnamestart Winn\surnameend} \&
  \bibinfo{author}{A.~\surnamestart Zisserman\surnameend}:
  \emph{\bibinfo{title}{The {PASCAL} {V}isual {O}bject {C}lasses {C}hallenge
  2012 {(VOC2012)} {R}esults}}.
\newblock
  \bibinfo{howpublished}{http://www.pascal-network.org/challenges/VOC/voc2012/workshop/index.html}.

\bibitemdeclare{inproceedings}{8500497}
\bibitem{8500497}
\bibinfo{author}{Eduardo \surnamestart Fernandez-Moral\surnameend},
  \bibinfo{author}{Renato \surnamestart Martins\surnameend},
  \bibinfo{author}{Denis \surnamestart Wolf\surnameend} \&
  \bibinfo{author}{Patrick \surnamestart Rives\surnameend}
  (\bibinfo{year}{2018}): \emph{\bibinfo{title}{A New Metric for Evaluating
  Semantic Segmentation: Leveraging Global and Contour Accuracy}}.
\newblock In: {\slshape \bibinfo{booktitle}{2018 IEEE Intelligent Vehicles
  Symposium (IV)}}, pp. \bibinfo{pages}{1051--1056},
  \doi{10.1109/IVS.2018.8500497}.

\bibitemdeclare{inproceedings}{DBLP:journals/corr/HeZRS15}
\bibitem{DBLP:journals/corr/HeZRS15}
\bibinfo{author}{Kaiming \surnamestart He\surnameend}, \bibinfo{author}{Xiangyu
  \surnamestart Zhang\surnameend}, \bibinfo{author}{Shaoqing \surnamestart
  Ren\surnameend} \& \bibinfo{author}{Jian \surnamestart Sun\surnameend}
  (\bibinfo{year}{2016}): \emph{\bibinfo{title}{Deep Residual Learning for
  Image Recognition}}.
\newblock In: {\slshape \bibinfo{booktitle}{2016 IEEE Conference on Computer
  Vision and Pattern Recognition (CVPR)}}, pp. \bibinfo{pages}{770--778},
  \doi{10.1109/CVPR.2016.90}.

\bibitemdeclare{article}{hinton2015distilling}
\bibitem{hinton2015distilling}
\bibinfo{author}{Geoffrey \surnamestart Hinton\surnameend},
  \bibinfo{author}{Oriol \surnamestart Vinyals\surnameend} \&
  \bibinfo{author}{Jeff \surnamestart Dean\surnameend} (\bibinfo{year}{2015}):
  \emph{\bibinfo{title}{Distilling the Knowledge in a Neural Network}}.
\newblock \doi{10.48550/ARXIV.1503.02531}.
\newblock \urlprefix\url{https://arxiv.org/abs/1503.02531}.

\bibitemdeclare{incollection}{inam2018safety}
\bibitem{inam2018safety}
\bibinfo{author}{Rafia \surnamestart Inam\surnameend}, \bibinfo{author}{Elena
  \surnamestart Fersman\surnameend}, \bibinfo{author}{Klaus \surnamestart
  Raizer\surnameend}, \bibinfo{author}{Ricardo \surnamestart Souza\surnameend},
  \bibinfo{author}{Amadeu \surnamestart Nascimento\surnameend} \&
  \bibinfo{author}{Alberto \surnamestart Hata\surnameend}
  (\bibinfo{year}{2018}): \emph{\bibinfo{title}{Safety for Automated Warehouse
  exhibiting collaborative robots}}.
\newblock In: {\slshape \bibinfo{booktitle}{Safety and Reliability--Safe
  Societies in a Changing World}}, \bibinfo{publisher}{CRC Press}, pp.
  \bibinfo{pages}{2021--2028}, \doi{10.1201/9781351174664-254}.

\bibitemdeclare{inproceedings}{8502466}
\bibitem{8502466}
\bibinfo{author}{Rafia \surnamestart Inam\surnameend}, \bibinfo{author}{Klaus
  \surnamestart Raizer\surnameend}, \bibinfo{author}{Alberto \surnamestart
  Hata\surnameend}, \bibinfo{author}{Ricardo \surnamestart Souza\surnameend},
  \bibinfo{author}{Elena \surnamestart Forsman\surnameend},
  \bibinfo{author}{Enyu \surnamestart Cao\surnameend} \&
  \bibinfo{author}{Shaolei \surnamestart Wang\surnameend}
  (\bibinfo{year}{2018}): \emph{\bibinfo{title}{Risk Assessment for Human-Robot
  Collaboration in an automated warehouse scenario}}.
\newblock In: {\slshape \bibinfo{booktitle}{2018 IEEE 23rd International
  Conference on Emerging Technologies and Factory Automation (ETFA)}},
  \bibinfo{volume}{1}, pp. \bibinfo{pages}{743--751},
  \doi{10.1109/ETFA.2018.8502466}.

\bibitemdeclare{article}{krizhevsky2009learning}
\bibitem{krizhevsky2009learning}
\bibinfo{author}{Alex \surnamestart Krizhevsky\surnameend},
  \bibinfo{author}{Geoffrey \surnamestart Hinton\surnameend} et~al.
  (\bibinfo{year}{2009}): \emph{\bibinfo{title}{Learning multiple layers of
  features from tiny images}}.

\bibitemdeclare{inproceedings}{lin2014microsoft}
\bibitem{lin2014microsoft}
\bibinfo{author}{Tsung-Yi \surnamestart Lin\surnameend},
  \bibinfo{author}{Michael \surnamestart Maire\surnameend},
  \bibinfo{author}{Serge \surnamestart Belongie\surnameend},
  \bibinfo{author}{James \surnamestart Hays\surnameend},
  \bibinfo{author}{Pietro \surnamestart Perona\surnameend},
  \bibinfo{author}{Deva \surnamestart Ramanan\surnameend},
  \bibinfo{author}{Piotr \surnamestart Doll{\'a}r\surnameend} \&
  \bibinfo{author}{C.~Lawrence \surnamestart Zitnick\surnameend}
  (\bibinfo{year}{2014}): \emph{\bibinfo{title}{Microsoft COCO: Common Objects
  in Context}}.
\newblock In \bibinfo{editor}{David \surnamestart Fleet\surnameend},
  \bibinfo{editor}{Tomas \surnamestart Pajdla\surnameend},
  \bibinfo{editor}{Bernt \surnamestart Schiele\surnameend} \&
  \bibinfo{editor}{Tinne \surnamestart Tuytelaars\surnameend}, editors:
  {\slshape \bibinfo{booktitle}{Computer Vision -- ECCV 2014}},
  \bibinfo{publisher}{Springer International Publishing},
  \bibinfo{address}{Cham}, pp. \bibinfo{pages}{740--755},
  \doi{10.1007/978-3-319-10602-1\_48}.

\bibitemdeclare{inproceedings}{marchisio2019deep}
\bibitem{marchisio2019deep}
\bibinfo{author}{Alberto \surnamestart Marchisio\surnameend},
  \bibinfo{author}{Muhammad~Abdullah \surnamestart Hanif\surnameend},
  \bibinfo{author}{Faiq \surnamestart Khalid\surnameend},
  \bibinfo{author}{George \surnamestart Plastiras\surnameend},
  \bibinfo{author}{Christos \surnamestart Kyrkou\surnameend},
  \bibinfo{author}{Theocharis \surnamestart Theocharides\surnameend} \&
  \bibinfo{author}{Muhammad \surnamestart Shafique\surnameend}
  (\bibinfo{year}{2019}): \emph{\bibinfo{title}{Deep learning for edge
  computing: Current trends, cross-layer optimizations, and open research
  challenges}}.
\newblock In: {\slshape \bibinfo{booktitle}{2019 IEEE Computer Society Annual
  Symposium on VLSI (ISVLSI)}}, \bibinfo{organization}{IEEE}, pp.
  \bibinfo{pages}{553--559}, \doi{10.1109/ISVLSI.2019.00105}.

\bibitemdeclare{inproceedings}{marchisio2018prunet}
\bibitem{marchisio2018prunet}
\bibinfo{author}{Alberto \surnamestart Marchisio\surnameend},
  \bibinfo{author}{Muhammad~Abdullah \surnamestart Hanif\surnameend},
  \bibinfo{author}{Maurizio \surnamestart Martina\surnameend} \&
  \bibinfo{author}{Muhammad \surnamestart Shafique\surnameend}
  (\bibinfo{year}{2018}): \emph{\bibinfo{title}{Prunet: Class-blind pruning
  method for deep neural networks}}.
\newblock In: {\slshape \bibinfo{booktitle}{2018 International Joint Conference
  on Neural Networks (IJCNN)}}, \bibinfo{organization}{IEEE}, pp.
  \bibinfo{pages}{1--8}, \doi{10.1109/IJCNN.2018.8489764}.

\bibitemdeclare{article}{matsubara2020head}
\bibitem{matsubara2020head}
\bibinfo{author}{Yoshitomo \surnamestart Matsubara\surnameend},
  \bibinfo{author}{Davide \surnamestart Callegaro\surnameend},
  \bibinfo{author}{Sabur \surnamestart Baidya\surnameend},
  \bibinfo{author}{Marco \surnamestart Levorato\surnameend} \&
  \bibinfo{author}{Sameer \surnamestart Singh\surnameend}
  (\bibinfo{year}{2020}): \emph{\bibinfo{title}{Head Network Distillation:
  Splitting Distilled Deep Neural Networks for Resource-Constrained Edge
  Computing Systems}}.
\newblock {\slshape \bibinfo{journal}{IEEE Access}} \bibinfo{volume}{8}, pp.
  \bibinfo{pages}{212177--212193}, \doi{10.1109/ACCESS.2020.3039714}.

\bibitemdeclare{article}{matsubara2020neural}
\bibitem{matsubara2020neural}
\bibinfo{author}{Yoshitomo \surnamestart Matsubara\surnameend} \&
  \bibinfo{author}{Marco \surnamestart Levorato\surnameend}
  (\bibinfo{year}{2020}): \emph{\bibinfo{title}{Neural Compression and
  Filtering for Edge-assisted Real-time Object Detection in Challenged
  Networks}}.
\newblock {\slshape \bibinfo{journal}{CoRR}} \bibinfo{volume}{abs/2007.15818},
  \doi{10.48550/arXiv.2007.15818}.
\newblock \eprint{2007.15818}.

\bibitemdeclare{article}{molchanov2016pruning}
\bibitem{molchanov2016pruning}
\bibinfo{author}{Pavlo \surnamestart Molchanov\surnameend},
  \bibinfo{author}{Stephen \surnamestart Tyree\surnameend},
  \bibinfo{author}{Tero \surnamestart Karras\surnameend}, \bibinfo{author}{Timo
  \surnamestart Aila\surnameend} \& \bibinfo{author}{Jan \surnamestart
  Kautz\surnameend} (\bibinfo{year}{2016}): \emph{\bibinfo{title}{Pruning
  Convolutional Neural Networks for Resource Efficient Transfer Learning}}.
\newblock {\slshape \bibinfo{journal}{CoRR}} \bibinfo{volume}{abs/1611.06440},
  \doi{10.48550/arXiv.1611.06440}.
\newblock \eprint{1611.06440}.

\bibitemdeclare{article}{paszke2017automatic}
\bibitem{paszke2017automatic}
\bibinfo{author}{Adam \surnamestart Paszke\surnameend}, \bibinfo{author}{Sam
  \surnamestart Gross\surnameend}, \bibinfo{author}{Soumith \surnamestart
  Chintala\surnameend}, \bibinfo{author}{Gregory \surnamestart
  Chanan\surnameend}, \bibinfo{author}{Edward \surnamestart Yang\surnameend},
  \bibinfo{author}{Zachary \surnamestart DeVito\surnameend},
  \bibinfo{author}{Zeming \surnamestart Lin\surnameend}, \bibinfo{author}{Alban
  \surnamestart Desmaison\surnameend}, \bibinfo{author}{Luca \surnamestart
  Antiga\surnameend} \& \bibinfo{author}{Adam \surnamestart Lerer\surnameend}
  (\bibinfo{year}{2017}): \emph{\bibinfo{title}{Automatic differentiation in
  pytorch}}.

\bibitemdeclare{inproceedings}{yu2015handwritten}
\bibitem{yu2015handwritten}
\bibinfo{author}{Naigong \surnamestart Yu\surnameend}, \bibinfo{author}{Panna
  \surnamestart Jiao\surnameend} \& \bibinfo{author}{Yuling \surnamestart
  Zheng\surnameend} (\bibinfo{year}{2015}): \emph{\bibinfo{title}{Handwritten
  digits recognition base on improved LeNet5}}.
\newblock In: {\slshape \bibinfo{booktitle}{The 27th Chinese Control and
  Decision Conference (2015 CCDC)}}, \bibinfo{organization}{IEEE}, pp.
  \bibinfo{pages}{4871--4875}, \doi{10.1109/CCDC.2015.7162796}.

\end{thebibliography}
\end{document}